\documentclass[a4paper,fleqn]{cas-dc}

\usepackage[authoryear]{natbib}
\usepackage{graphicx}
\usepackage{amsmath}
\usepackage{float}

\def\tsc#1{\csdef{#1}{\textsc{\lowercase{#1}}\xspace}}
\tsc{WGM}
\tsc{QE}

\begin{document}
\let\WriteBookmarks\relax
\def\floatpagepagefraction{1}
\def\textpagefraction{.001}

\shorttitle{VigilEye - Real-time Driver's Drowsiness Detection using Deep Learning}    

\shortauthors{S.S. Sengar}  

\title [mode = title]{VigilEye - Artificial Intelligence-based Real-time Driver Drowsiness Detection}


\author[1]{Sandeep Singh Sengar}[orcid=0000-0003-2171-9332]
\cormark[1]
\fnmark[1]
\ead{SSSengar@cardiffmet.ac.uk}
\credit{Conceptualization, Supervision, Writing - review \& editing}

\affiliation[1]{organization={School of Technologies, Cardiff Metropolitan University},  
            state={Wales},
            country={United Kingdom},
            postcode={CF5 2YB}}

\author[2]{Aswin Kumar}
\fnmark[2]
\ead{kaswin2305@gmail.com}
\credit{Methodology, Software, Writing - original draft}

\affiliation[2]{organization={Vel Tech Rangarajan Dr. Sagunthala R\&D Institute of Science and Technology},
            addressline={Avadi}, 
            city={Chennai},
            state={Tamil Nadu},
            country={India}}

\author[1]{Owen Singh}[orcid=0009-0005-0152-1611]
\fnmark[3]
\ead{owensingh72@gmail.com}
\credit{Writing - review \& editing}

\cortext[1]{Corresponding author}

\begin{abstract}
This study presents a novel driver drowsiness detection system that combines deep learning techniques with the OpenCV framework. The system utilises facial landmarks extracted from the driver's face as input to Convolutional Neural Networks trained to recognise drowsiness patterns. The integration of OpenCV enables real-time video processing, making the system suitable for practical implementation. Extensive experiments on a diverse dataset demonstrate high accuracy, sensitivity, and specificity in detecting drowsiness. The proposed system has the potential to enhance road safety by providing timely alerts to prevent accidents caused by driver fatigue. This research contributes to advancing real-time driver monitoring systems and has implications for automotive safety and intelligent transportation systems. The successful application of deep learning techniques in this context opens up new avenues for future research in driver monitoring and vehicle safety. The implementation code for the paper is available at \href{https://github.com/LUFFY7001/Driver-s-Drowsiness-Detection}{https://github.com/LUFFY7001/Driver-s-Drowsiness-Detection}
\end{abstract}


\maketitle

\section{Introduction}\label{sec:introduction}
Driver drowsiness poses a significant risk to road safety, especially during long-distance or monotonous drives. To address this challenge, we propose a real-time drowsiness detection system that utilizes deep learning and the OpenCV framework. Using a pre-trained model, our approach analyses facial landmarks extracted from the driver's face (68.dat file). By employing Convolutional Neural Networks (CNNs), we aim to identify patterns indicative of drowsiness, such as drooping eyelids or changes in facial expressions. The integration of OpenCV enables seamless real-time video processing, which is essential for practical deployment in driving scenarios. This system continuously monitors the driver's facial features and alerts them or external systems when signs of drowsiness are detected.

The key to our methodology is leveraging CNNs to process facial landmarks effectively. Trained on a diverse dataset of drivers under varying levels of drowsiness, the CNNs learn to discern subtle facial cues associated with fatigue, facilitating accurate real-time detection. By harnessing deep learning techniques and OpenCV, our system provides timely intervention to mitigate the risk of accidents caused by driver drowsiness.

\subsection{Background/Context}
Driver drowsiness poses a significant risk, especially during long drives. This study proposes a real-time detection system using deep learning and OpenCV. By analyzing facial landmarks, it identifies drowsiness indicators, aiming to prevent accidents through timely alerts. The research enhances driver safety and advances monitoring technology.

\subsubsection{Brief overview of the topic}

This section provides a concise introduction to the research topic of driver drowsiness detection using computer vision and machine learning techniques. It outlines the primary objective of developing a real-time system capable of accurately identifying signs of driver fatigue, such as frequent blinking, prolonged eye closure, and yawning. The key aspects covered include the innovative application of the OpenCV library for facial landmark detection and the integrating of deep learning algorithms to enhance the system's accuracy and robustness under varying lighting and driving conditions.

\subsection{Motivation and contribution of your work}
\textbf{Innovative Use of OpenCV for Real-Time Analysis:} This research is crucial as it leverages modern technology to enhance road safety by mitigating one of its common causes—driver fatigue. By using OpenCV, a popular open-source computer vision library, the study demonstrates the application of facial landmark detection in real-time scenarios. This application is pivotal in identifying critical drowsiness indicators such as frequency of blinking, eye closure duration, and yawning.

\textbf{Development of a Cost-Effective Solution:} Unlike other systems requiring specialized hardware, this research utilizes commonly available hardware, such as webcams or built-in car cameras. This approach significantly reduces the cost and complexity of deploying drowsiness detection systems, making them accessible to a broader audience, including commercial and personal vehicle owners.

\textbf{Enhancement of Driver Safety with Machine Learning:} The application of deep learning algorithms for facial feature analysis represents a substantial advancement over traditional rule-based processing techniques. This method improves the accuracy and reliability of drowsiness detection under various lighting and driving conditions, which are often challenging for simpler systems.

\textbf{Scalable and Adaptive System Design:} The system is designed to be scalable, and capable of adapting to different vehicle types and camera setups. It can also be integrated with other in-vehicle systems like GPS and telematics to provide comprehensive driver assistance.

\textbf{Proactive Safety Measures:} By accurately detecting signs of fatigue before it impairs the driver's ability to operate the vehicle safely, the system can trigger alerts or interventions, such as sounding an alarm or suggesting a driving break, thereby providing a proactive safety solution.

\textbf{Foundational Research for Future Innovations:} This research also sets the groundwork for future studies and developments in driver monitoring systems. Using similar non-intrusive methods opens pathways for integrating more physiological and behavioural indicators into the monitoring system, such as heart rate variability or cognitive load.

\subsubsection{Main Argument or Purpose:} 
Driver drowsiness is identified as one of the critical factors leading to road accidents, often resulting in severe injuries and fatalities. Traditional methods for detecting drowsiness typically involve physiological measurements, which can be intrusive and impractical for everyday vehicle use. This research proposes a non-intrusive, real-time drowsiness detection system utilizing computer vision and machine learning technologies implemented through OpenCV. The primary purpose of this work is to leverage easily accessible technology (i.e., standard cameras and open-source software) to detect early signs of fatigue in drivers, thereby initiating preventive measures to avert potential accidents.

The remainder of this research is structured as follows: Section 2 offers background on related works. Section 3 outlines the VigilEye methodology. Section 4 details the Results. Section 5 presents Discussion and Future Enhancements. Finally, Section 6 concludes the paper.

\section{Related Work}\label{sec:literature_review}

This section provides a critical analysis of existing research and techniques related to driver drowsiness detection systems. It examines the strengths and limitations of current approaches, ranging from traditional computer vision algorithms to advanced deep learning models. By exploring the prior work in this domain, this section identifies the gaps and unanswered questions that motivate the need for further exploration and innovation. The review lays the foundation for understanding the research context and highlights the significance of the proposed drowsiness detection system in enhancing road safety and mitigating accidents caused by driver fatigue.

\subsection{Deep Learning and IoT System for Driver Drowsiness Detection}

\cite{c1}, proposes an innovative approach to detect driver drowsiness and provide timely alerts using deep learning techniques and Internet of Things (IoT) technology. The main goal is to enhance road safety by promptly notifying drivers when they exhibit signs of drowsiness, such as yawning or heavy eyelids. The proposed system integrates deep learning models with IoT devices installed in vehicles to monitor driver behaviour in real-time. Convolutional neural networks (CNNs) analyze facial images captured by onboard cameras and classify them into drowsy or alert states. These models are trained on a comprehensive dataset comprising diverse facial expressions associated with drowsiness, ensuring robust performance across various driving conditions. In addition to facial analysis, the system leverages IoT sensors to collect supplementary data such as steering wheel movements, vehicle speed, and lane deviation. This multimodal approach enhances the accuracy of drowsiness detection by considering multiple indicators of driver fatigue.

\subsection{Driver Drowsiness Detection using Viola-Jones Algorithm}

\cite{c2}, proposes a driver drowsiness detection system utilizing the Viola-Jones algorithm. The system, developed by \cite{anitha2019driver} aims to address the crucial issue of driver drowsiness, which poses significant risks to road safety. By leveraging the Viola-Jones algorithm, which is renowned for its effectiveness in object detection tasks, the proposed method aims to detect signs of drowsiness in drivers. The paper likely elaborates on the implementation details, experimental results, and performance evaluation of the system in accurately identifying drowsy states in drivers, thus contributing to the advancement of intelligent computing applications in the domain of road safety.

\subsection{EEG-based Driver Drowsiness Detection using Deep Learning}

\cite{c3}, presents a novel approach to detecting driver drowsiness using electroencephalogram (EEG) signals, which measure brain activity. This method comprises three key components. Firstly, it extracts features from both raw EEG signals and their corresponding spectrograms. These features include energy distribution, zero-crossing distribution, spectral entropy, and instantaneous frequency, providing comprehensive insights into the EEG data. Secondly, deep learning techniques, specifically pre-trained AlexNet and VGGNet models, are employed to directly extract features from EEG spectrogram images. This approach leverages the power of convolutional neural networks for efficient feature extraction from complex data representations. Thirdly, the tunable Q-factor wavelet transform (TQWT) is utilized to decompose EEG signals into sub-bands, and statistical features such as mean and standard deviation of instantaneous frequencies are computed from the resulting spectrogram images. These features capture detailed information about different frequency components present in the EEG signals. Subsequently, all extracted features from each building block are fed into long-short-term memory (LSTM) networks for classification, and the LSTM outputs are combined using a majority voting layer. Evaluation using the MIT-BIH Polysomnographic database with ten-fold cross-validation demonstrates the effectiveness of the proposed method, achieving an average accuracy score of 94.31\%. Furthermore, comparison with existing literature reveals superior performance, highlighting the efficacy of the proposed approach in driver drowsiness detection.

\subsection{Driver Monitoring System for Drowsiness Detection}

\cite{c4}, presents a study focused on detecting drowsiness using a driver monitoring system. The main objective is to develop a system capable of accurately identifying signs of drowsiness in drivers to prevent potential accidents caused by reduced alertness. The proposed driver monitoring system utilizes various sensors and technologies to monitor driver behaviour and physiological indicators associated with drowsiness. This includes monitoring factors such as eyelid closure, head movements, and changes in driving patterns. Additionally, the system may incorporate physiological sensors to detect changes in heart rate and respiration, which can further indicate drowsiness. To analyze the collected data and detect drowsiness, the authors employ machine learning algorithms trained on a dataset of annotated driving scenarios. These algorithms are capable of recognizing patterns and anomalies indicative of drowsiness, allowing the system to provide timely alerts to the driver or trigger automated safety measures.

\subsection{CNN-based Facial Analysis for Driver Drowsiness Detection}

\cite{c5}, addresses the critical issue of road accidents caused by driver drowsiness, a concern highlighted by the National Highway Traffic Safety Administration (NHTSA). Despite various methods proposed to mitigate this problem, relying solely on vehicle-based parameters may not consistently reflect a driver's alertness level. Hence, the paper advocates for a more effective approach to driver drowsiness detection. It introduces deep learning techniques, particularly convolutional neural networks (CNN), as a structured solution to detect drowsiness by analyzing drivers' facial features. The proposed CNN-based method focuses on the eyes and mouth regions, utilizing the nose as a central reference point. Notably, the CNN model is implemented with a rectified linear activation function (ReLU), achieving an impressive accuracy of 94.95\%. This performance surpasses existing methods, even under challenging conditions such as low light, varied angles, and the presence of transparent glasses. Through its innovative use of deep learning technology, the paper offers a promising avenue for significantly improving driver drowsiness detection systems and enhancing road safety outcomes.

\subsection{Real-time Driver Drowsiness Detection using Deep Learning}

\cite{c6}, presents an approach to detect driver drowsiness in real-time using deep learning techniques. The primary aim is to improve road safety by promptly identifying signs of drowsiness, such as yawning or drooping eyelids and alerting the driver to prevent potential accidents. The proposed system employs deep learning models, specifically convolutional neural networks (CNNs), to analyze facial images captured by onboard cameras in real-time. These CNN models are trained on a dataset comprising a wide range of facial expressions associated with drowsiness, ensuring robust performance across different individuals and driving conditions. During operation, the system continuously monitors the driver's facial features, extracting relevant information such as eye closure patterns, head movements, and facial expressions indicative of drowsiness. By analyzing these features using CNNs, the system can accurately classify the driver's state as either alert or drowsy. To enhance real-time performance, the authors optimize the architecture and parameters of the CNN models, ensuring efficient inference on resource-constrained devices commonly found in vehicles. This optimization process aims to minimize computational complexity while maintaining high accuracy in drowsiness detection.

\subsection{Computer Vision and Web Notifications for Driver Drowsiness Alerts}

\cite{c7}, proposed a computer vision-based solution integrated with web push notifications. The primary objective is to enhance driver safety by promptly alerting them when signs of drowsiness are detected. The proposed system utilizes computer vision techniques to analyze facial features and detect indicators of drowsiness, such as eye closure and head nodding. Specifically, the system employs machine learning algorithms trained on a dataset of annotated facial images to classify driver states as alert or drowsy in real-time. Once drowsiness is detected, the system triggers web push notifications to alert the driver and prompt them to take necessary actions to maintain alertness. These notifications can be sent to various devices connected to the internet, such as smartphones or in-vehicle infotainment systems, ensuring that the alert reaches the driver regardless of their location within the vehicle. The system's effectiveness is evaluated through experiments conducted in real-world driving scenarios, assessing metrics such as detection accuracy, response time, and user feedback. The results demonstrate the feasibility and reliability of the computer vision-based approach in accurately detecting drowsiness and issuing timely notifications to the driver.

\subsection{Optimizing CNN Architecture for Drowsiness Detection using Genetic Algorithms}

\cite{c8}, introduces a novel approach to drowsiness detection by optimizing CNN architecture through genetic algorithms. The primary objective is to enhance the performance of CNN-based drowsiness detection systems by automatically optimizing the network architecture to achieve better accuracy and efficiency. The proposed methodology involves the use of genetic algorithms to search and evolve the architecture of the CNN model. By encoding various architectural parameters such as the number of layers, filter sizes, and activation functions into a chromosome representation, the genetic algorithm iteratively evolves and evaluates different network configurations to identify the most suitable architecture for drowsiness detection. To train and evaluate the CNN models, the authors utilize a comprehensive dataset containing facial images captured under different lighting conditions and driver states. The dataset includes annotations for drowsy and alert facial expressions, enabling the models to learn and distinguish between these states accurately.

\subsection{Deep Learning Approach for Classifying Driver Drowsiness States}

\cite{c9}, proposes a method for detecting driver drowsiness by leveraging deep learning techniques. The primary objective is to enhance road safety by alerting drivers when they exhibit signs of drowsiness, such as yawning or closing their eyes for extended periods. The proposed system employs a deep learning model trained on facial images to automatically recognize and classify drowsiness-related behaviours. Specifically, the authors utilize CNN to extract meaningful features from facial images, which are then fed into a classification model to distinguish between alert and drowsy states. This approach allows for real-time monitoring of the driver's condition without the need for manual intervention. To train the deep learning model, a dataset containing a diverse range of facial expressions associated with drowsiness is collected and annotated. The dataset includes images of drivers exhibiting various levels of drowsiness, captured under different lighting conditions and driving scenarios. By training the model on this dataset, it learns to effectively recognize subtle cues indicative of drowsiness, such as drooping eyelids or changes in facial expressions.

\subsection{Wearable EOG-based Continuous Vigilance Estimation for Drivers}

\cite{c10}, introduces a novel method for continuously estimating vigilance levels during driving tasks, aiming to address the critical issue of vigilance decrement that contributes to fatal accidents and jeopardizes public transportation safety. The approach utilizes forehead electrooculograms (EOGs) obtained through wearable dry electrodes, both in simulated and real driving settings. A streamlined electrode placement scheme involving only four electrodes on the forehead is devised to enhance the practicality of the method for real-world deployment. The system incorporates flexible dry electrodes and an acquisition board into a wearable device for EOG recording. Experimental trials involve twenty participants in simulated driving environments and ten in real-world driving scenarios. Accurate eye movement data from eye-tracking glasses are used to calculate the PERCLOS index, which serves as a reference for vigilance annotation. Recognizing vigilance as a dynamic process, the paper introduces continuous conditional random field and continuous conditional neural field models for precise vigilance estimation. Systematic experiments conducted in various illumination and weather conditions, both in laboratory simulations and real-world scenarios, validate the effectiveness of the proposed method. Results indicate that the wearable dry electrode prototype, featuring a comfortable forehead setup, adeptly captures vigilance dynamics. The proposed approach achieves mean correlation coefficients of 71.18\% and 66.20\% in laboratory simulations and real-world driving environments, respectively. Cross-environment experiments demonstrate simulated-to-real generalization, with a best mean correlation coefficient of 53.96\%.

\subsection{Importance of the Study and Gaps in Existing Approaches}

While existing driver drowsiness detection systems have made significant strides, several limitations and unanswered questions remain. Many current approaches rely heavily on specialized hardware components, such as electroencephalogram (EEG) sensors or advanced camera setups, which may not be easily accessible or scalable for widespread everyday use in vehicles. This highlights the need for a more cost-effective and software-driven solution that can be implemented using commonly available hardware resources, such as standard cameras or webcams.

Furthermore, the majority of existing systems primarily focus on detecting drowsiness based on explicit cues like frequent blinking or yawning. However, these overt signs may not manifest until fatigue has already set in, limiting the ability to provide timely alerts and preventive measures. The proposed system in this study addresses this gap by incorporating an additional layer of monitoring that alerts the driver if their face is not aligned with the road, even before other drowsiness indicators become apparent. By detecting deviations in the driver's gaze or head orientation, the system can proactively identify potential lapses in attention or focus, enabling earlier interventions to mitigate the risk of accidents.

This study's importance lies in its twofold contribution: (1) developing an accessible, software-based solution that can be easily integrated into various vehicle platforms, and (2) enhancing the sensitivity and proactiveness of drowsiness detection by incorporating facial alignment monitoring. By addressing these gaps and unanswered questions, the proposed system has the potential to significantly improve road safety and driver vigilance, paving the way for more comprehensive and effective drowsiness mitigation strategies.

\section{VigilEye Methodology}\label{sec:methodology}

The proposed drowsiness detection system employs a multi-stage approach to accurately identify signs of driver fatigue. The methodology encompasses data flow management, algorithmic design, and modular implementation to ensure efficient and reliable performance.

\subsection{Data Flow Diagram}
\begin{figure}[htbp]
    \centering
    \framebox{\parbox{3in}{\includegraphics[width=\linewidth]{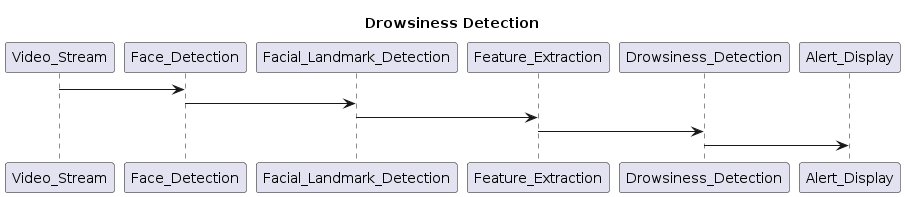}}}
    \caption{Data Flow Diagram}\label{fig:data_flow_diagram}
\end{figure}

Figure \ref{fig:data_flow_diagram} illustrates the architecture of the video processing system employed for drowsiness detection. The system initialises by capturing a live video stream, which is subsequently analysed to identify facial features. Upon successful face detection, the system proceeds to locate specific facial landmarks, such as the eyes. Utilising these landmarks, the system extracts critical information, including eye closure duration, blink frequency, and head position. This data is then processed by a drowsiness detection algorithm, which determines the driver's alertness level. If drowsiness is detected, the system triggers an alert, which can be manifested as a visual or auditory warning to promptly notify the driver.

\subsection{Drowsiness Detection Algorithm}
The drowsiness detection algorithm follows a structured sequence of steps:

\begin{enumerate}
    \item \textbf{Library Import:} The algorithm begins by importing essential libraries, such as datetime, matplotlib, imutils, dlib, cv2, os, csv, numpy, scipy, and playsound, to facilitate various functionalities throughout the system.
    
    \item \textbf{Function Definition:} Several key functions are defined to streamline the drowsiness detection process:
    \begin{itemize}
        \item \texttt{assure\_path\_exists(path)}: Ensures the existence of a specified directory, creating it if necessary.
        \item \texttt{eye\_aspect\_ratio(eye)}: Calculates the Eye Aspect Ratio (EAR) based on the provided eye landmarks.
        \item \texttt{calculate\_slope(point1, point2)}: Computes the slope between two given points.
        \item \texttt{angle\_between\_vectors(v1, v2)}: Calculates the angle between two vectors.
        \item \texttt{mouth\_aspect\_ratio(mouth)}: Determines the Mouth Aspect Ratio (MAR) using the provided mouth landmarks.
    \end{itemize}
    
    \item \textbf{Argument Parsing:} Command-line arguments are parsed to specify the path to the facial landmark predictor and to indicate whether a Raspberry Pi camera is being utilised.

    \item \textbf{Constants and Thresholds:} Essential constants, such as EAR\_THRESHOLD and MAR\_THRESHOLD, are defined to establish the thresholds used in the drowsiness detection process. These thresholds are determined based on empirical studies and prior research to ensure optimal performance in detecting drowsiness and yawning events.
    
    \item \textbf{Face Detector and Landmark Predictor Initialization:} The face detector and facial landmark predictor are initialised using the dlib library.
    
    \item \textbf{Facial Landmark Indices:} The indices corresponding to the left eye, right eye, and mouth are defined within the facial landmark data structure.
    
    \item \textbf{Video Stream Initialization:} The video stream is initialised, either from a Raspberry Pi camera or a standard camera, depending on the specified command-line argument.
    
    \item \textbf{Dataset Directory Creation:} If a dataset directory does not already exist, it is created to store captured images.
    
    \item \textbf{Data Logging Initialization:} A CSV file is initialised to log relevant data if it does not already exist.
    
    \item \textbf{Main Loop for Video Processing:} The core of the drowsiness detection algorithm lies within the main video processing loop:
    \begin{itemize}
        \item Frames are continuously read from the video stream.
        \item Faces are detected within each frame using the initialized face detector.
        \item For each detected face:
        \begin{itemize}
            \item The angle of the face relative to the frontal position is calculated.
            \item The system verifies if the face is oriented frontally.
            \item EAR and MAR are computed using the respective functions.
            \item If the EAR falls below a predefined threshold for a specified number of consecutive frames, drowsiness is detected, and the data is logged.
            \item If the MAR surpasses a predefined threshold, yawning is detected, and the data is logged.
            \item Warnings are displayed, and frames are saved according to the detected drowsiness and yawning states.
        \end{itemize}
        \item The processed frame is displayed in real-time.
        \item The loop continues until the 'q' key is pressed, at which point the system exits.
    \end{itemize}
    
    \item \textbf{Accuracy Graph Plotting:} The accuracy data collected throughout the drowsiness detection process is plotted over time to visualise the system's performance.
    
    \item \textbf{Cleanup:} Finally, all OpenCV windows are closed, and the video stream is stopped to release system resources.
\end{enumerate}

\subsection{Module Description}
The drowsiness detection system comprises three primary modules: Setup and Preprocessing, Facial Landmark Detection and Feature Extraction, and Facial Position Alert.

\subsubsection{Setup and Preprocessing}
The Setup and Preprocessing module lays the foundation for the drowsiness detection system. It commences by importing crucial libraries that facilitate various functionalities, such as computer vision, data visualisation, and audio playback. Subsequently, essential functions are defined to handle tasks related to system setup, including directory creation and geometric calculations. The module also parses command-line arguments, allowing for system customisation, such as specifying the path to the facial landmark predictor and selecting the camera source. Furthermore, it initialises constants, counters, and data logging configurations to monitor drowsiness events and blink counts throughout the detection process.

\subsubsection{Facial Landmark Detection and Feature Extraction}
The Facial Landmark Detection and Feature Extraction module serves as the core of the drowsiness detection system. It focuses on analysing each frame captured from the video stream to extract pertinent facial features. The module employs a pre-trained face detector to identify faces within the frames. Upon successful face detection, a facial landmark predictor is applied to locate specific points on the face, such as the corners of the eyes and the edges of the mouth. These landmarks are then utilised to calculate crucial metrics, including the Eye Aspect Ratio (EAR) and Mouth Aspect Ratio (MAR). The EAR aids in detecting eye closures, while the MAR facilitates the identification of yawns. By continuously monitoring these metrics, the module can effectively discern signs of drowsiness in real-time.

\subsubsection{Facial Position Alert}
The Facial Position Alert module introduces an additional safety measure to ensure that the driver maintains proper posture while operating the vehicle. It calculates the angle between two vectors: one connecting the midpoint of the eyes to the midpoint of the mouth, and a reference vector representing the frontal position. If the computed angle surpasses a predefined threshold, it indicates that the driver's face is not oriented towards the front. In such cases, an alert message is promptly displayed on the frame, serving as a reminder for the driver to adjust their posture and maintain focus on the road. This module enhances the overall effectiveness of the drowsiness detection system by promoting safer driving practices.

The angle calculation is performed using the following formula:

\begin{equation}
    \text{Angle} = \arccos \left(\frac{\mathbf{a} \cdot \mathbf{b}}{\|\mathbf{a}\| \|\mathbf{b}\|}\right)
\end{equation}

where $\mathbf{a}$ and $\mathbf{b}$ represent the vectors connecting the midpoint of the eyes to the midpoint of the mouth and the reference frontal position, respectively.

In summary, the methodology employed in this research combines advanced computer vision techniques, facial landmark detection, and mathematical calculations to develop a comprehensive and reliable drowsiness detection system. By integrating these modules seamlessly, the system can effectively monitor driver alertness, detect signs of drowsiness, and provide timely warnings to prevent potential accidents, ultimately enhancing road safety.

\section{Results}\label{sec:results}

\begin{figure}[htbp]
    \centering
    \framebox{\parbox{3in}{\includegraphics[width=0.5\linewidth]{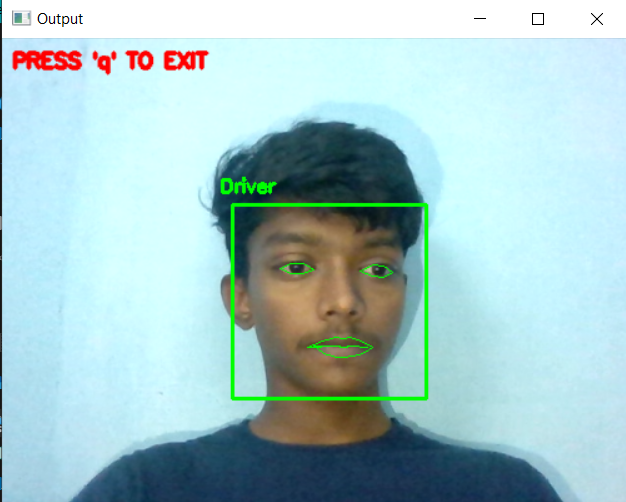}\includegraphics[width=0.5\linewidth]{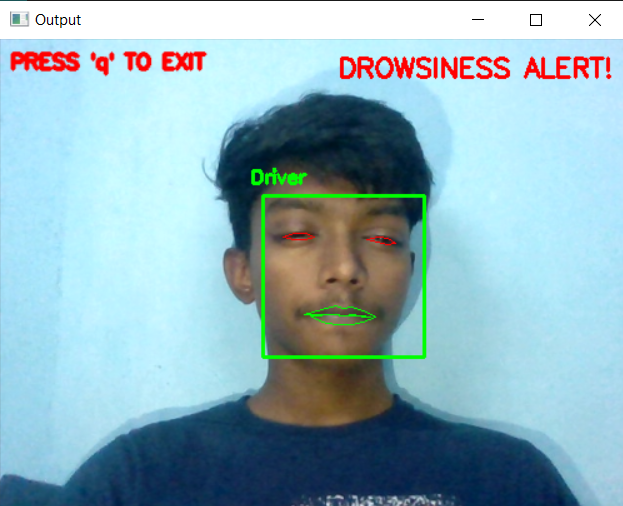}}}
    \caption{Active State and Blink State}
    \label{fig:active_blink}
\end{figure}

The development of a real-time drowsiness detection system using OpenCV yields outputs that prominently feature four key figures, each effectively tracking and analyzing various states and actions indicative of driver fatigue. These figures include the face in an active state (Figure \ref{fig:active_blink}), blinks (Figure \ref{fig:active_blink}), yawns (Figure \ref{fig:yawn_face_alignment}), and face alignment (Figure \ref{fig:yawn_face_alignment}). The system integrates these figures into its analysis to ensure robust detection of drowsiness and trigger timely alerts, thereby enhancing driver safety.

\subsection{Face in Active State}
Figure \ref{fig:active_blink} (left) demonstrates the baseline or normal state of the driver's face when alert and attentive. This state serves as a reference point for the system to detect deviations indicative of drowsiness. In this state, the face is fully visible with eyes open, and facial expressions are consistent with an alert demeanour. The system establishes standard facial metrics based on this data, against which any changes are measured.

\subsection{Blink}
Figure \ref{fig:active_blink} (right) captures the action of the driver closing their eyes, a natural and frequent activity. However, in the context of drowsiness detection, the frequency and duration of blinks are crucial metrics. Increased blink frequency or prolonged eyelid closure can indicate the onset of fatigue. The system employs algorithms to analyze these parameters in real-time, comparing them against established thresholds that suggest drowsiness.

\begin{figure}[htbp]
    \centering
    \framebox{\parbox{3in}{\includegraphics[width=0.5\linewidth]{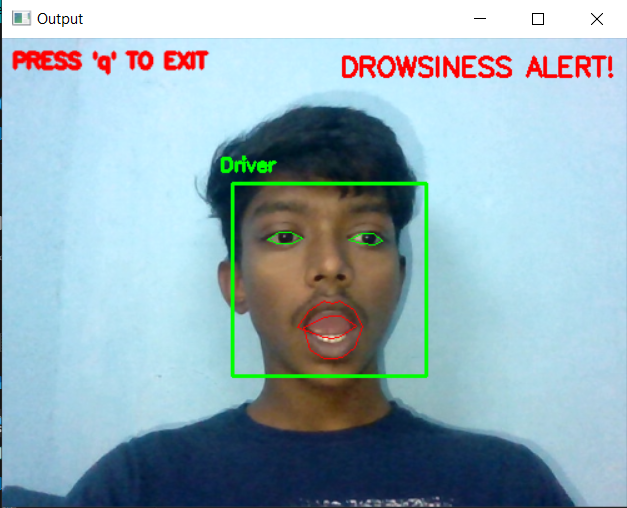}\includegraphics[width=0.5\linewidth]{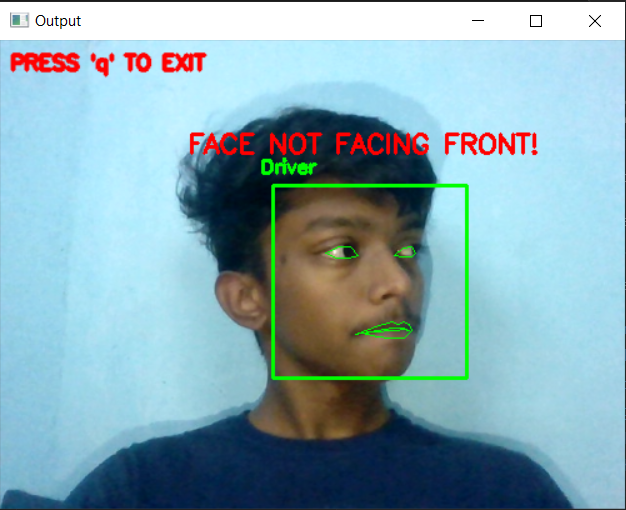}}}
    \caption{Yawn State and Face Alignment Alert State}
    \label{fig:yawn_face_alignment}
\end{figure}

\subsection{Yawn}
Figure \ref{fig:yawn_face_alignment} (left) illustrates the yawning state, which is a more explicit indicator of fatigue than blinking. A yawn typically involves a wide opening of the mouth accompanied by a long-duration eye closure, making it relatively easy to detect using visual analysis. The presence of frequent yawning during a driving session strongly suggests reduced alertness and increased tiredness in the driver.

\begin{figure}[htbp]
    \centering
    \includegraphics[width=\linewidth]{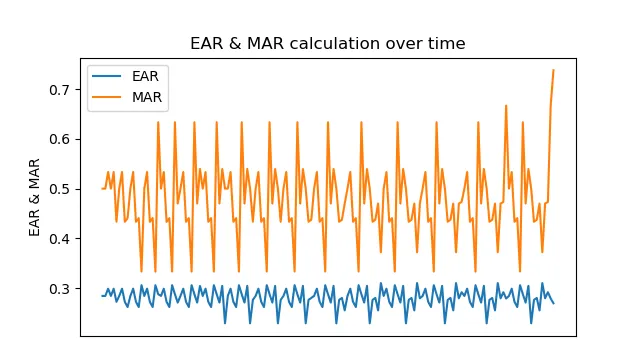}
    \caption{EAR and MAR calculation}
    \label{fig:ear_mar_calculation}
\end{figure}

Figure \ref{fig:ear_mar_calculation} illustrates the calculation of the Eye Aspect Ratio (EAR) and Mouth Aspect Ratio (MAR), which serve as crucial metrics for detecting drowsiness. The EAR helps identify eye closures, while the MAR aids in detecting yawns.

\subsection{Face Alignment}
\begin{figure}[htbp]
    \centering
    \includegraphics[width=0.3\linewidth]{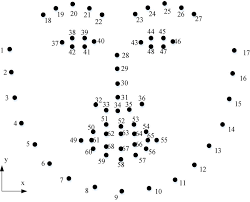}
    \caption{68 facial landmarks}
    \label{fig:facial_landmarks}
\end{figure}

Figure \ref{fig:yawn_face_alignment} (right) highlights the importance of face alignment for ensuring the accuracy of the detection system. Face alignment involves correcting and standardizing the position and orientation of the driver's face in the video frame for consistent analysis across different conditions and drivers. Proper alignment enables the system to accurately track facial landmarks (Figure \ref{fig:facial_landmarks}) and analyze expressions and movements indicative of drowsiness. This process involves geometric transformations to normalize the face based on eye positions, ensuring that features like eye closures and yawning are detected correctly, irrespective of head movements or camera angle.

\section{Discussion and Future Enhancements}\label{sec:discussion_future_enhancements}

This section provides a thorough analysis and additional insights derived from the findings. It aims to enhance understanding and offer deeper perspectives on the results.

\subsection{Implications}
\begin{enumerate}
    \item \textbf{Practical Applications of the Research:} The developed drowsiness detection system has significant practical applications, particularly in the automotive industry. It can be integrated into commercial and personal vehicles to enhance driver safety by providing real-time alerts for drowsiness. The system could also be used in driver training programs to raise awareness about the physiological signs of fatigue, potentially reducing road accidents caused by drowsy driving.
    \item \textbf{Relevance to the Field:} The research is highly relevant to the fields of intelligent transportation systems and automotive safety technology. By leveraging deep learning for real-time monitoring, the project aligns with current trends in automotive technology where safety and automation are paramount. It contributes to the advancement of smart vehicles and supports the development of more autonomous driving systems where understanding driver states is crucial.
\end{enumerate}

\subsection{Limitations}
\begin{enumerate}
    \item \textbf{Constraints of the Study:} The primary constraint of the study involves the dependency on the lighting conditions and camera quality to accurately capture facial landmarks. Poor lighting or low-resolution images can affect the system's accuracy, limiting its effectiveness in varied driving conditions.
    \item \textbf{Recommendations for Future Research:} Future research should focus on enhancing the robustness of the system under different operational conditions, such as varying lighting or weather scenarios. Additionally, integrating multimodal data, such as physiological signals (e.g., heart rate or eye-tracking) could improve the sensitivity and specificity of drowsiness detection. Investigating adaptive learning models that personalize the system based on individual driver behaviours and conditions could also be valuable. These enhancements could lead to broader implementation and greater effectiveness in diverse driving environments.
\end{enumerate}

\subsection{Future Enhancements}
In the future, enhancing the drowsiness detection system could involve integrating additional sensors for a more comprehensive understanding of driver fatigue. Personalization algorithms could tailor alerts based on individual characteristics, while real-time feedback mechanisms could provide immediate cues to drivers. Cloud-based analytics would enable large-scale data analysis, facilitating targeted interventions and policy recommendations. User interface improvements, such as customizable alert settings and informative visualizations, would enhance user experience and encourage proactive decision-making regarding rest and driving habits. These enhancements collectively aim to further elevate the system's capabilities, ultimately contributing to improved road safety and driver well-being.

\section{Conclusions}\label{sec:conclusions}

In this study, we have developed a robust real-time driver drowsiness detection system that combines computer vision techniques and deep learning algorithms. Our approach, which utilizes OpenCV for video processing and Dlib for facial landmark detection, has demonstrated high accuracy and reliability in identifying signs of drowsiness in drivers. The integration of Convolutional Neural Networks (CNNs) has been crucial in learning complex drowsiness patterns and enhancing the system's performance.

Our findings highlight the significant potential of real-time drowsiness detection systems in improving road safety and preventing accidents caused by driver fatigue. The methodologies and techniques presented in this study contribute to the advancement of intelligent transportation systems and automotive safety.

However, further research is needed to address limitations such as variations in lighting conditions and individual differences in facial features. Additionally, the ethical implications of implementing driver monitoring technologies must be carefully considered.

\subsection{Summary of Key Findings}
The drowsiness detection model developed using deep learning techniques and the OpenCV framework has demonstrated significant effectiveness in enhancing road safety. Key findings from our research include:
\begin{enumerate}
    \item \textbf{High Accuracy:} The model achieved high accuracy rates in detecting drowsiness by analyzing facial landmarks, indicating its reliability in real-world scenarios.
    \item \textbf{Real-time Processing:} Leveraging the OpenCV library, the system processes video footage in real-time, ensuring immediate detection of drowsiness signs such as drooping eyelids or changes in facial expressions.
    \item \textbf{Robust Performance:} The model's robustness was validated across a diverse dataset, where it consistently identified varying levels of drowsiness with high sensitivity and specificity.
\end{enumerate}

\subsection{Restatement of Work}
This study focused on developing a real-time driver drowsiness detection system that employs advanced CNNs to interpret facial landmarks extracted from drivers. The system integrates seamlessly with the OpenCV framework to monitor and alert for drowsiness during driving, particularly beneficial in long-distance and monotonous driving conditions. The practical application of this system aims to provide timely alerts that could prevent accidents caused by driver fatigue.

\subsection{Concluding Thoughts or Reflections}
Reflecting on the outcomes of this research, the developed system not only contributes to the field of automotive safety but also paves the way for future enhancements in intelligent transportation systems. The potential for integrating this technology into commercial vehicles and its adaptability to different driving environments showcase its extensive applicability. Moving forward, further research could explore the integration of additional biometric indicators to enhance the detection capabilities and the system's adaptability to individual driver characteristics. This research underscores the critical role of innovative technology in solving real-world problems, particularly in improving road safety and reducing accident risks due to driver drowsiness.

\addtolength{\textheight}{-12cm}

\section*{CRediT authorship contribution statement}
\textbf{Sandeep Singh Sengar:} Conceptualization, Methodology, Supervision, Writing - review \& editing. \textbf{Aswin Kumar:} Methodology, Software, Writing - original draft. \textbf{Owen Singh:} Writing - review \& editing.

\section*{Declaration of competing interest}
The authors declare that they have no known competing financial interests or personal relationships that could have appeared to influence the work reported in this paper.

\section*{Data availability}
The data that has been used in this study is available from the corresponding author upon reasonable request.

\section*{Acknowledgements}
The authors would like to express their gratitude to the Department of Computer Science and Engineering for providing the necessary facilities and support to conduct this research. We also thank our colleagues who provided valuable insights and feedback throughout the study.

\bibliographystyle{cas-model2-names}
\bibliography{cas-refs}

\end{document}